\documentclass{article}

\usepackage[final, nonatbib]{nips_2018}

\usepackage[utf8]{inputenc} 
\usepackage[T1]{fontenc}    
\usepackage{hyperref}       
\usepackage{url}            
\usepackage{booktabs}       
\usepackage{amsfonts}       
\usepackage{nicefrac}       
\usepackage{microtype}      
 
\usepackage{graphicx}
\usepackage{color}
\usepackage[hang,flushmargin]{footmisc}
\usepackage{subcaption}
\usepackage{enumitem}

\title{Let's Play Again: Variability of Deep Reinforcement Learning Agents in Atari Environments}

\newcommand{\kaleigh}{    
  Kaleigh Clary \\
  University of Massachusetts Amherst\\
  Amherst, MA 01003 \\
  \texttt{kclary@cs.umass.edu}
}

\newcommand{\john}{
  John Foley \\
  Smith College \\
  Northampton, MA 01060 \\
  \texttt{jjfoley@smith.edu} 
}

\newcommand{\david}{
    David Jensen \\
  University of Massachusetts Amherst\\
  Amherst, MA 01003 \\
    \texttt{jensen@cs.umass.edu}    
}

\newcommand{\emma}{
Emma Tosch\\
  University of Massachusetts Amherst\\
  Amherst, MA 01003 \\
\texttt{etosch@cs.umass.edu}
}

\author{
\kaleigh{}\\
\And
\emma{}\\
  \And
\john{}\\
\And
\david{}
}

\begin{document}

\maketitle

\begin{abstract}
Reproducibility in reinforcement learning is challenging:  uncontrolled stochasticity from many sources, such as the learning algorithm, the learned policy, and the environment itself have led researchers to report the performance of learned agents using aggregate metrics of performance over multiple random seeds for a single environment. Unfortunately, there are still pernicious sources of variability in reinforcement learning agents that make reporting common summary statistics an unsound metric for performance. Our experiments demonstrate the variability of common agents used in the popular OpenAI Baselines repository. We make the case for reporting post-training agent performance as a distribution, rather than a point estimate.
\end{abstract}

\section{Introduction}
Deep reinforcement learning faces substantial and unusual challenges in evaluation and reproducibility~\cite{henderson2017deep, rlblogpost}. Based on reports of common evaluation practices in the field, many RL researchers train a few candidate models\footnote{We use the term model to refer to the learned component of a policy.} and then report performance using some aggregate function of the scores of the trained agents.\footnote{We use \textit{score} to mean any measure of performance used to compare the performance of different RL agents.} This aggregate can be taken from the learning curves of the model or from scores collected in post-training episodes.\footnote{Evaluation of online learning methods is beyond the scope of this work.} 
Such metrics are often used for comparison against reported benchmark results. 
Underlying this practice is the expectation that the average reward provides a meaningful summary of agent performance when operating in the environment. 

Researchers know that randomness in the training algorithm of a deep RL agent can have a large impact on the agent's ability to effectively learn a policy. For example, an agent with poorly seeded weights, or poorly chosen random actions early in training, may find its way to a local optimum and never achieve high reward. With this context in mind, deep RL researchers often train multiple agents with different random seeds to account for this variability in training.

Unfortunately, careful selection of random seeds during training does not necessarily translate to careful consideration of reporting metrics for reproducibility.
Reporting and analyzing the choice of random seed is critical for reproducing a training procedure, but it is also critical for assessing the variability inherent in the final evaluation of an agent. Reproducibility in RL is increasingly a cause for concern, due to the growing complexity of training, evaluation, and agent architecture. The process of selecting and reporting random seeds represents a potential bias in the training of RL agents and requires closer inspection, as there is high potential for researchers to unknowingly run afoul of the \emph{multiple comparisons procedure}. 

To our knowledge, this paper is the first to explore variability of individual deep RL models. We will use the term \emph{seeded model} to refer to the specific model specified by hyperparameters and model random seed. We constrain our analysis to models operating in Atari environments~\cite{bellemare13arcade} and leave a broader analysis to future work.

We show empirically that seeded models have diverse performance distributions, demonstrate how this can affect model selection, and point to commonly used techniques for correcting for variability.

\section{Problems with Reproducibility and Reporting Practices}

There has been much recent interest in improving reproducibility in deep reinforcement learning~\cite{henderson2017deep,rlblogpost,islam2017reproducibility}. 
One major challenge to reproducibility is the effect of uncontrolled variability during either training or evaluation. 
Recent work has attempted to isolate the effects of various sources of variability in RL, examining variability due to differences in algorithm implementation, hyperparameter selection, environment stochasticity, network architectures, and random seed selection~\cite{henderson2017deep}.
We focus on reproducibility and variability as it pertains to the effects of erroneously reporting point estimates, and specifically analyze the implications of  the effects of random seeds. A common practice is to extract a single sample of performance from each seeded model from the learning curve of that model and report it as representative of a learned agent's behavior~\cite{cohen2018distributed, henderson2017deep}.
 However, without characterizing the variability of a seeded model and and specifying how the single model was selected, it is not clear that a single sample is sufficient to describe the behavior of that model. 

\textbf{Model selection and random seeds.} 
Most commonly, researchers select one model from among many, based on which model has the highest \textit{mean} score across some number of trials (each associated with a different random seed)~\cite{baselines,henderson2017deep,Mnih2016}. Less commonly, researchers select one model from among many based on which model has the highest \textit{maximum} score across some number of trials~\cite{Mnih2016, NIPS2017_7112}.

Either approach can produce unexpected biases due to the high variability of scores.  Using the mean score is an appropriate performance measure if the score distribution is well-behaved (e.g., Gaussian). However, if the performance distribution is multimodal, has fat tails, or has outliers, the mean alone can be a misleading summary of model performance. Using the maximum score can be extremely misleading if the different models produce score distributions with widely differing variance.  In such a case, choosing based on the maximum score will favor models with high-variance distributions rather than models with highest expected value~\cite{Jensen2000}. There also exists a second-order effect in both cases, because the expected distribution of either the mean or the maximum has variance itself, and selecting models based on a sample from that distribution (a specific mean or maximum) will tend to favor those models with high inherent variability.

Aggregates of model performance summarizing behavior across random seeds can also be affected by variability of seeded model performance. With no convergence guarantees, less control over the representation of state, and the sources of variability listed previously, it is unclear whether one should expect individual trained deep RL agents to exhibit well-behaved, narrow distributions of performance. Reporting over a sample of seeds, each potentially exhibiting variability, introduces an additional degree of freedom in the statistical analysis of model performance. Thus, it is important to characterize the variability of agent performance associated with different random seeds.

\textbf{Background: Multiple Comparisons Procedures.} Jensen \& Cohen~\cite{Jensen2000} define a \textit{multiple comparison procedure} (MCP) as any process that generates multiple items (e.g., models), estimates a score (with some variability) for each item, and then selects the item with the maximum score.  MCPs are common in machine learning, and they result in positive bias in the score of the selected item.  This statistical effect of MCPs is the underlying reason for regularization methods such as complexity penalties and evaluation procedures such as cross-validation.

Failing to account for variability in model scores can lead to incorrect conclusions about model rankings. 
By selecting the models with the maximum score, it is likely that the maximum will be an outlier rather than an unbiased estimate of performance.
Selection procedures using the mean are also MCPs, as that mean score has variability (and possibly different variability depending on training approach, agent architecture, etc.).  The bias of such estimates is magnified as the number of models increases, as the length or number of testing episodes decreases, and as the inherent variability of performance increases.  Worse, if the inherent variability of models differs, then an MCP will often result in selecting the model with the highest variance, rather than the model with the highest expected score. 

Such high variability of results makes performance evaluations vulnerable to cherry-picked point estimates of performance and even scrupulous researchers can run into the problems outlined above. This can result in incorrect conclusions when comparing learning methods and architectures, and it can misdirect individual and field-wide threads of research~\cite{gelman2013garden}.

\section{Complications due to Variability}

We collected end-game scores of 100 games for each of 10 different seeded models and examined the resulting score distributions of each seeded model set.\footnote{Our code is available at: \url{https://github.com/kclary/variability-RL}.}

\paragraph{OpenAI Baselines Benchmarks.}
To create a suite of models, we replicate the experiments in OpenAI Baselines Benchmarks~\cite{baselines}. We train different models on 10 random seeds each for five ALE Atari environments\footnote{Researchers often insert randomly repeated actions to make Atari environments more stochastic. Our experiments use the ALE \textit{NoFrameskip-v4} Atari environments, which do not include these stochastic variations~\cite{bellemare13arcade}.}: Beam Rider, Breakout, Q*bert, Seaquest, and Space Invaders. We compare a2c~\cite{Mnih2016}, acktr~\cite{NIPS2017_7112}, and ppo2~\cite{schulman2017proximal} models trained using OpenAI Baselines implementations with default hyperparameters. Learning curves for our models are provided in Appendix A.

\paragraph{Variability of Seeded Models.}

\begin{figure*}[t!]
    \centering
    \begin{subfigure}[b]{0.48\textwidth}
    \includegraphics[width=0.95\textwidth]{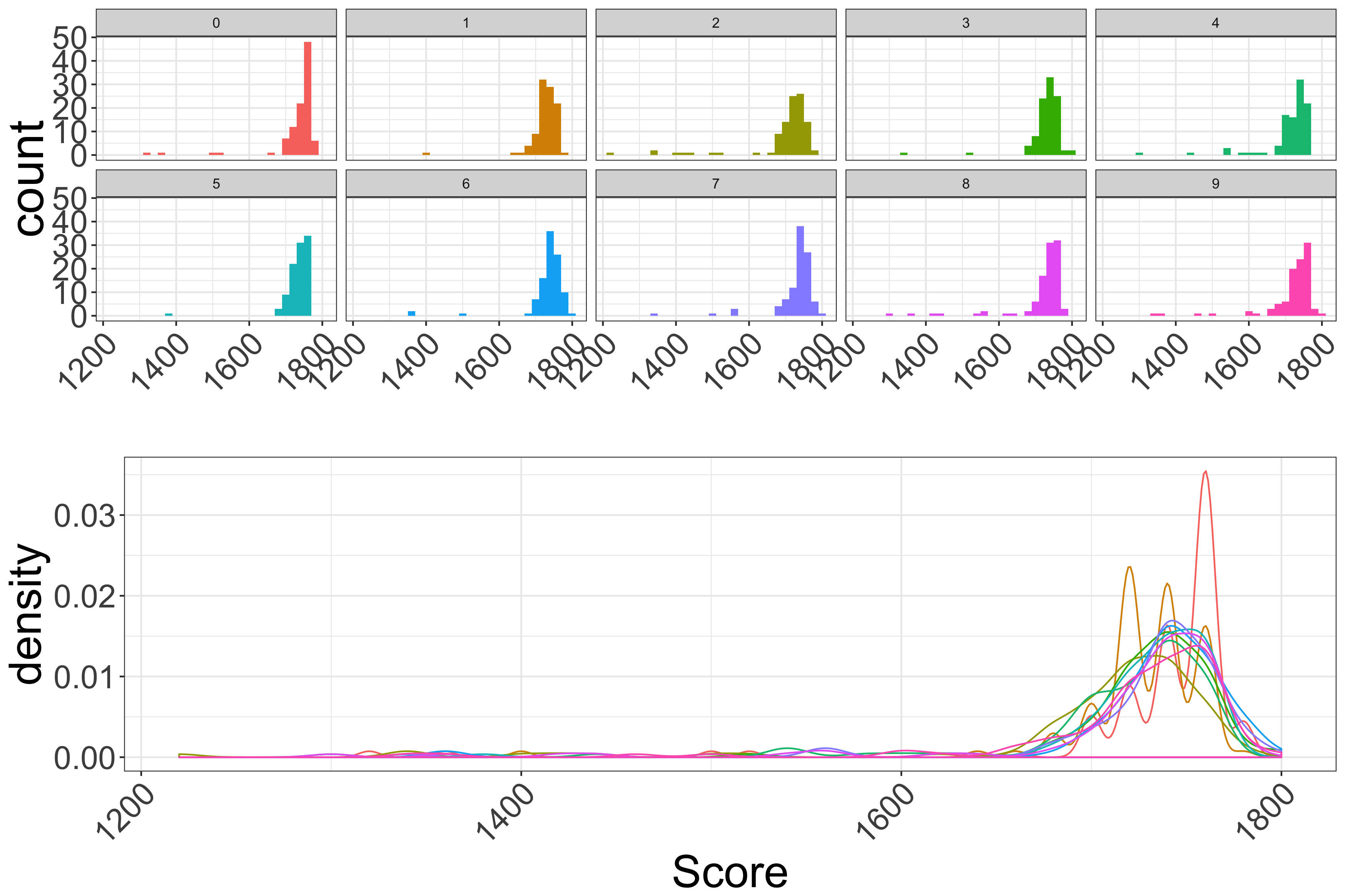}
    \caption{\label{fig:uniform}acktr on Seaquest}
    \end{subfigure}%
    ~\begin{subfigure}[b]{0.48\textwidth}
     \includegraphics[width=0.95\textwidth]{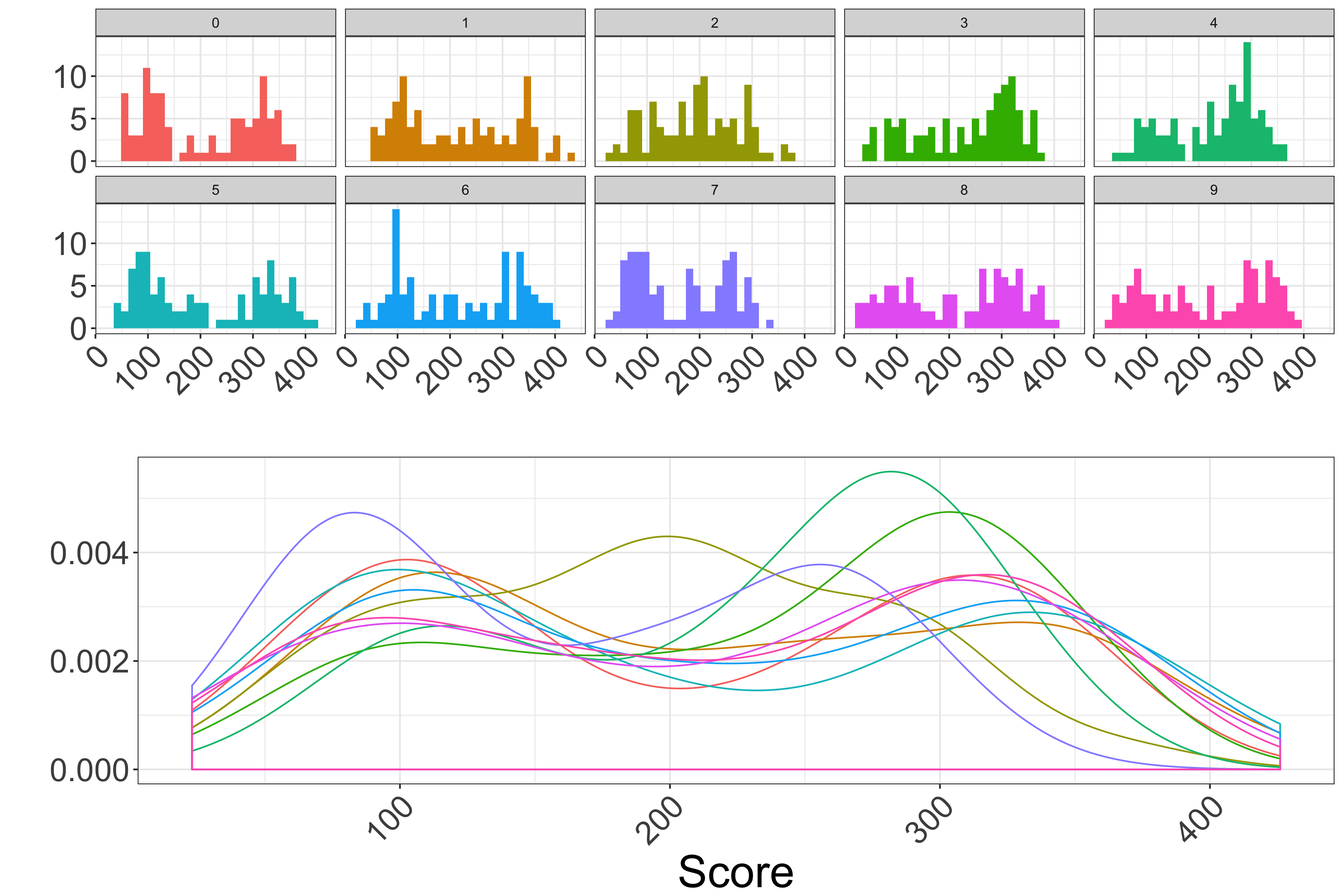}
    \caption{\label{fig:spread}ppo2 on Breakout}
    \end{subfigure}
    \begin{subfigure}[b]{0.48\textwidth}
     \includegraphics[width=0.95\textwidth]{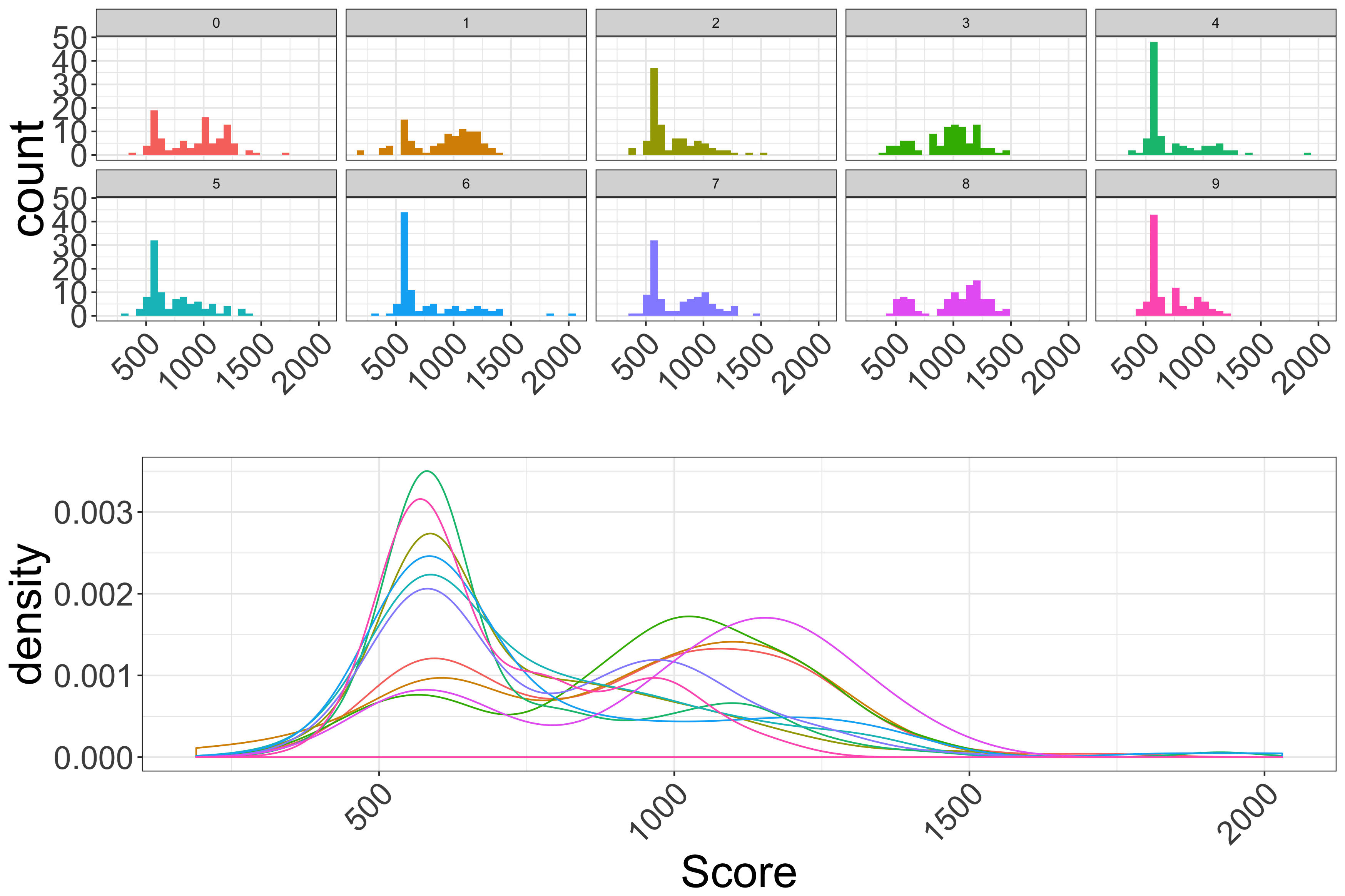}
     \caption{\label{fig:multimodal-space-invaders}ppo2 on Space Invaders}
     \end{subfigure}%
     ~\begin{subfigure}[b]{0.48\textwidth}
    \includegraphics[width=0.95\textwidth]{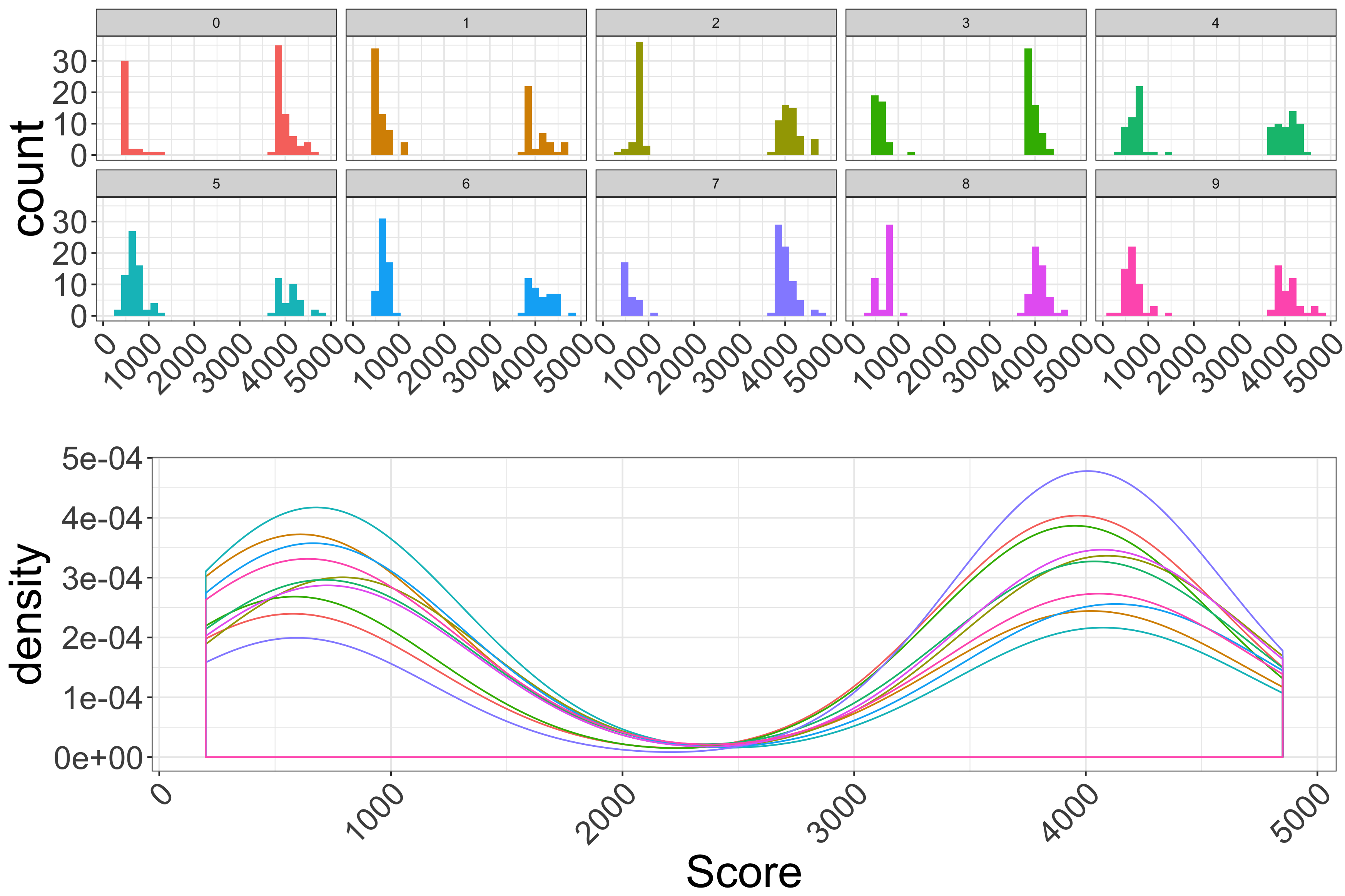}
    \caption{\label{fig:multimodal-qbert}a2c on Q*bert}
    \end{subfigure}
    \caption{\label{fig:curves} A selection of seeded models (each seed represented as a different color) on a subset of Atari games featured in the OpenAI Baselines benchmarks. Each agent-environment combination depicts histograms from 10 seeds. Each histogram represents the frequencies of scores for 100 trials of the given seed. Below the histograms are kernel density estimates for the distributions of scores. Some distributions have greater variability between seeds than others.}
\end{figure*}

We report score distributions for four model sets from our suite in Figure \ref{fig:curves}. These results illustrate why it is best practice to test and account for seeded model variability in random seed and model selection. 

\begin{itemize}[leftmargin=*]
\item\textbf{Stationary Performance Distributions across Random Seeds.} In Figure \ref{fig:uniform}, we see that each random seed exhibits similar performance distributions. Sampling a small number of times from each of these random seeds may look like variability between random seeds, when in fact the distribution of scores is due to variability of the seeded model.

\item\textbf{Fat-tailed Distributions.} The score distribution for Breakout models shown in Figure \ref{fig:spread} covers nearly the entire range of possible scores in the first level of Breakout. Reporting the mean, standard error of 209.3 $\pm$ 3.2 gives an inaccurate basis of comparison between Breakout models.

\item\textbf{Multimodal Distributions.}
The distribution of Q*bert scores shown in \ref{fig:multimodal-qbert} is bimodal\footnote{The modality exhibited by these models is not replicated in Q*bert agents trained with acktr or ppo2, so this distribution is likely not a consequence of the environment. For all three distribution plots, see Appendix~\ref{app:histograms}.} in seeded model performance, while Space Invaders scores (Figure \ref{fig:multimodal-space-invaders}) are bimodal when taking together the performance of all seeded models. This distinction is lost if we do not examine variability both within and between seeded models.
\end{itemize}

\paragraph{Degenerate Model Selection.} The mean for a2c on Q*bert (2425.4 $\pm$ 53.8) is not a reasonable expectation for model performance; it is among the least densely observed scores. The wide distribution of scores in ppo2 on Breakout means that the mean is not very informative. Using the max to choose models in any of these model sets does not guarantee a high-performing model. This means that not only is reporting difficult, but even selecting a good model for production use requires knowledge of within-model variance.

\section{Discussion}

These results show that seeded models can also exhibit variability in Atari deep RL. In light of this, we argue for reporting model performance as a distribution in place of point estimates. Even the mean with standard error is inappropriate when the distribution is bimodal. Note that generating the distribution of scores for a seeded model simply requires running the trained model for some number of trial episodes. This is far less resource intensive than training a larger number of models on new random seeds. To more accurately characterize the expected performance of models trained with particular hyperparameter settings, we suggest reporting the performance distributions of a few different random seeds to demonstrate a range of score performance across two axes of variability for a given set of hyperparameters.

To choose a model with expected maximum score among several seeded models for production or publication, use best practices and avoid reporting cherry-picked scores of outlier random seeds. Multiple comparisons procedures are a well-studied process in statistics, and there are several ways to correct for this variability when making a selection. Resampling scores from the model with highest score, Bonferroni adjustment of score samples, and cross-validation over partitions of model scores are each methods used to account for variability in MCPs. For a more complete discussion of these and other  MCP adjustment methods, see Jensen \& Cohen \cite{Jensen2000}.

Difficulty in finding random seeds and hyperparameters that seed reasonable model performance in deep RL has sometimes been described as \textit{instability}. Referring to these phenomena as instability implies that this behavior is somehow errant and unexpected. 
Our experiments imply that, at least in Atari environments, variability in seeded model performance is expected behavior, and commonly reported evaluation measures are insufficient for characterizing that variability. Using community standard implementations of algorithms and the deep networks that back them, we find that even seeded models operating in controlled environments can still exhibit wild variability in performance.    

We demonstrate the variability of seeded model performance in Atari environments. This variability impacts the conclusions one might draw from current reporting and evaluation practices. We recommend (1) running trained agents in the Atari environment for many episodes to collect samples of end-game score, (2) reporting this distribution of scores to characterize model performance, and (3) using statistical methods to adjust for seeded model variability when choosing among candidate models. 


\section*{Acknowledgements}
This material is based upon work supported by the United States Air Force under Contract No, FA8750-17-C-0120.  Any opinions, findings and conclusions or recommendations expressed in  this material are those of the author(s) and do not necessarily reflect the views of the United States Air Force.

\small
\bibliographystyle{plain}

\appendix

\newpage
\section{Learning Curves}
\label{app:learning_curves}

\begin{figure}[h!]
\centering
\begin{subfigure}[b]{0.6\textwidth}
\includegraphics[width=0.95\textwidth]{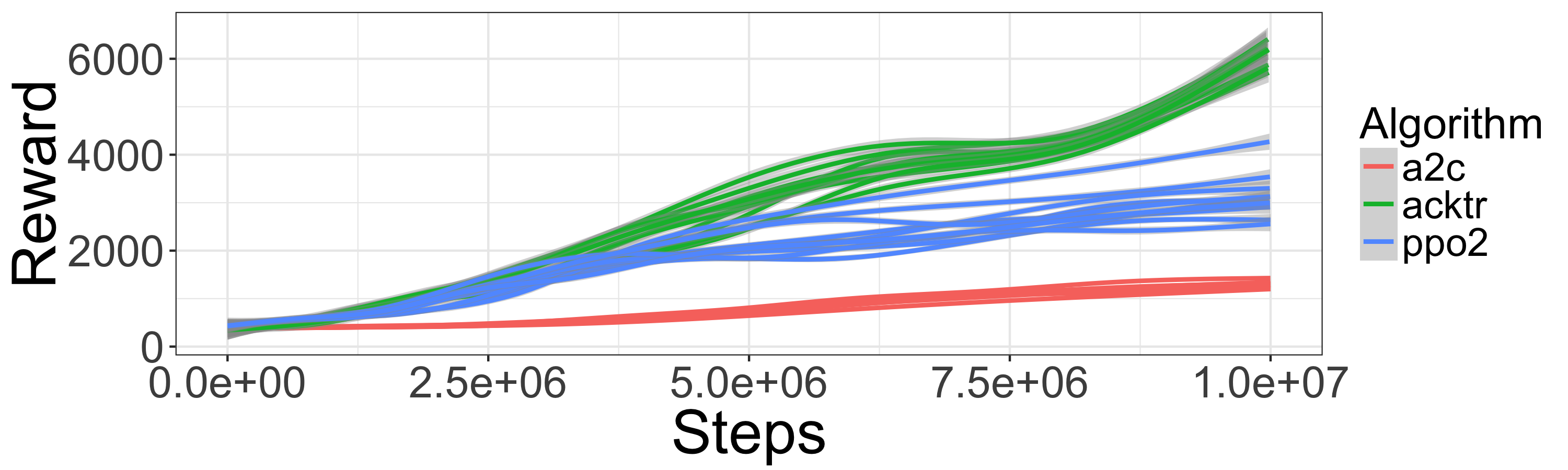}
\caption{\label{fig:beam-rider-lc} BeamRider}
\end{subfigure}
\begin{subfigure}[b]{0.6\textwidth}
\includegraphics[width=0.95\textwidth]{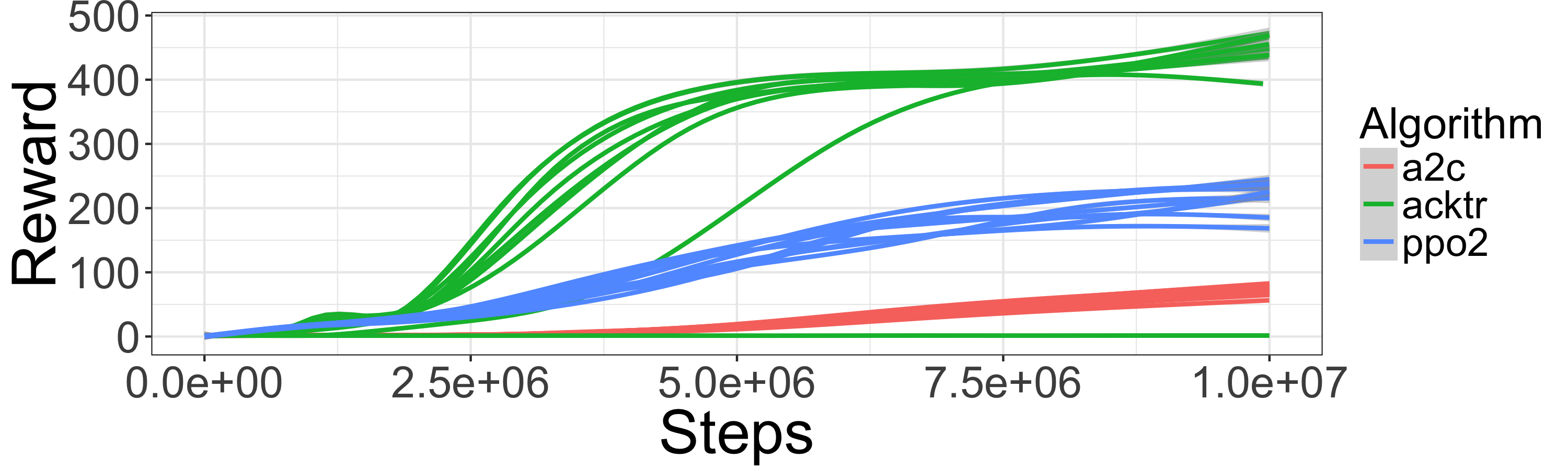}
\caption{\label{fig:breakout-lc} Breakout}
\end{subfigure}
\begin{subfigure}[b]{0.6\textwidth}
\includegraphics[width=0.95\textwidth]{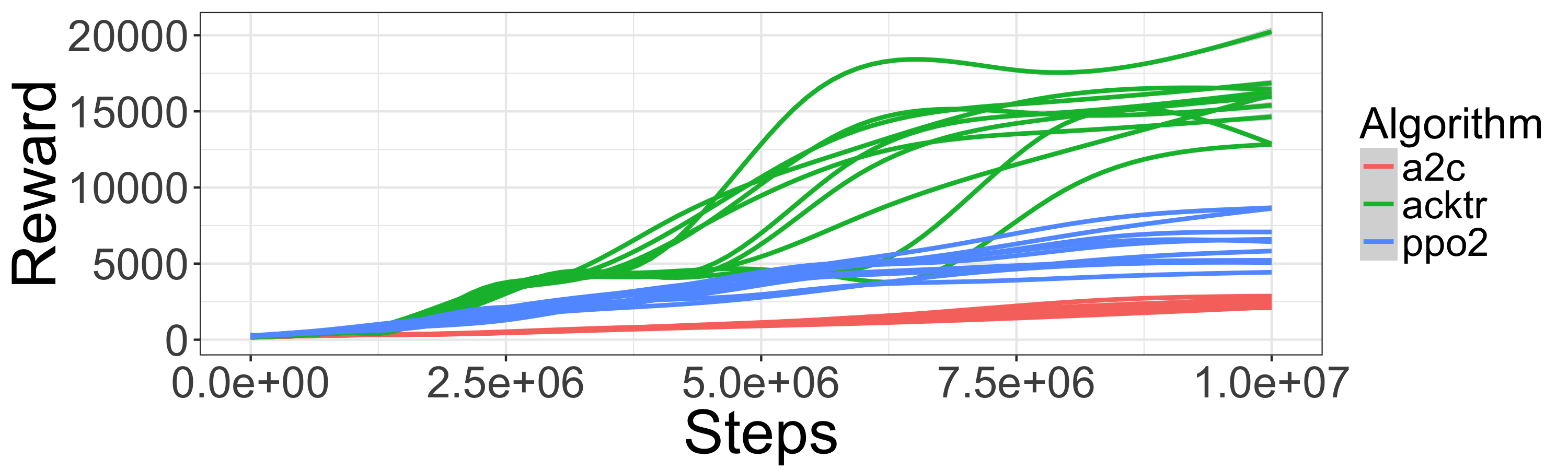}
\caption{\label{fig:qbert-lc} Q*Bert}
\end{subfigure}
\begin{subfigure}[b]{0.6\textwidth}
\includegraphics[width=0.95\textwidth]{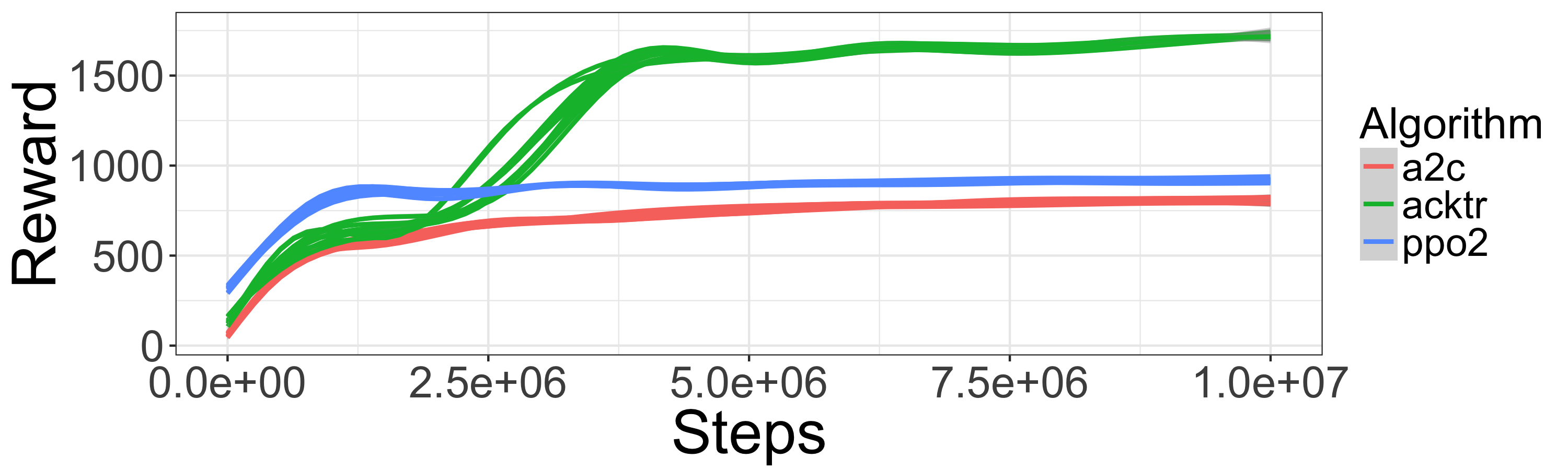}
\caption{\label{fig:seaquest-lc} Seaquest}
\end{subfigure}
\begin{subfigure}[b]{0.6\textwidth}
\includegraphics[width=0.95\textwidth]{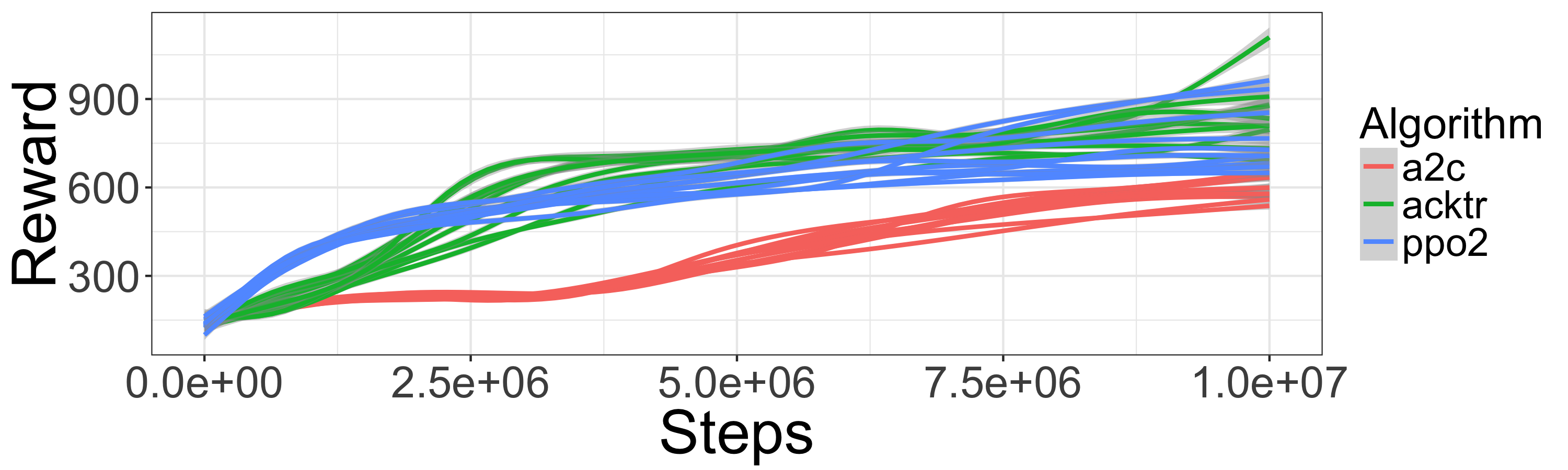}
\caption{\label{fig:space-invaders-lc} SpaceInvaders}
\end{subfigure}
\caption{\label{fig:complete-learning-curves} Learning curves for each agent (a2c~\cite{Mnih2016}, acktr~\cite{NIPS2017_7112}, ppo2~\cite{schulman2017proximal}) on each of the five Atari game environments featured in the OpenAI Baselines benchmarks.}
\end{figure}

\newpage

\section{Complete Agent-Environment Histograms}
\label{app:histograms}

The following five environments are featured in OpenAI Baseline's Atari benchmark (BeamRider, Breakout, Q*Bert, Seaquest, and SpaceInvaders). We chose three agents for evaluation: a2c~\cite{Mnih2016}, acktr~\cite{NIPS2017_7112}, and ppo2~\cite{schulman2017proximal}. The OpenAI benchmark includes six agents; we decided against deepq due to the required training time, acer due to a bug that was present in the Baselines code at the time of evaluation, and trpo\_mpi due to early issues during training related to MPI calls. 

Each agent-environment combination depicts the histograms from 10 seeds. Each histogram represents the frequencies of scores for 100 trials of the given seed. Below the histograms are kernel density estimates for the distributions of scores. 
    Some distributions have greater variability between seeds than others.

\begin{figure}[h!]
    \centering
    \begin{subfigure}[b]{0.33\textwidth}
    \includegraphics[width=0.99\textwidth]{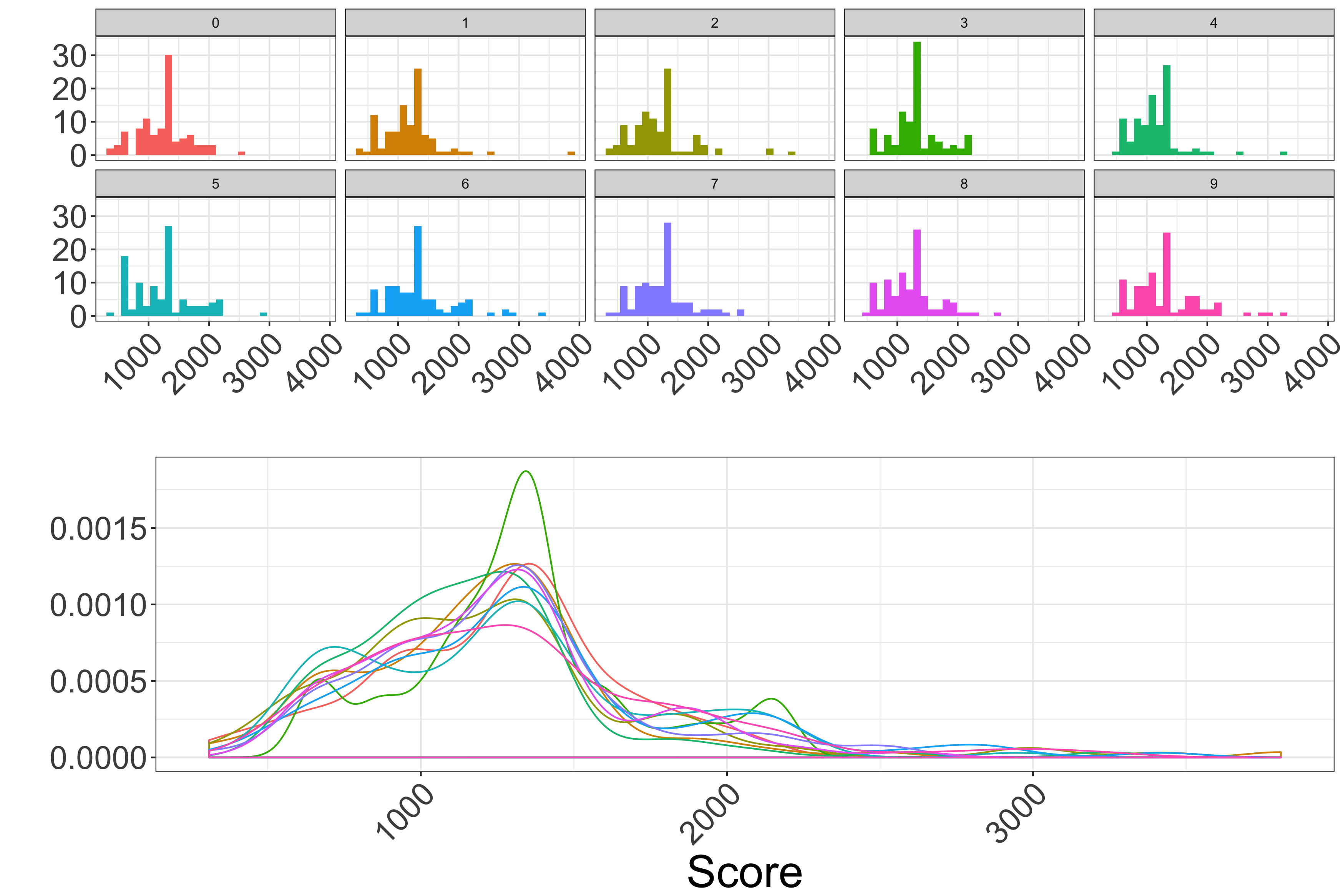}
    \caption{a2c}
    \end{subfigure}%
    \begin{subfigure}[b]{0.33\textwidth}
    \includegraphics[width=0.99\textwidth]{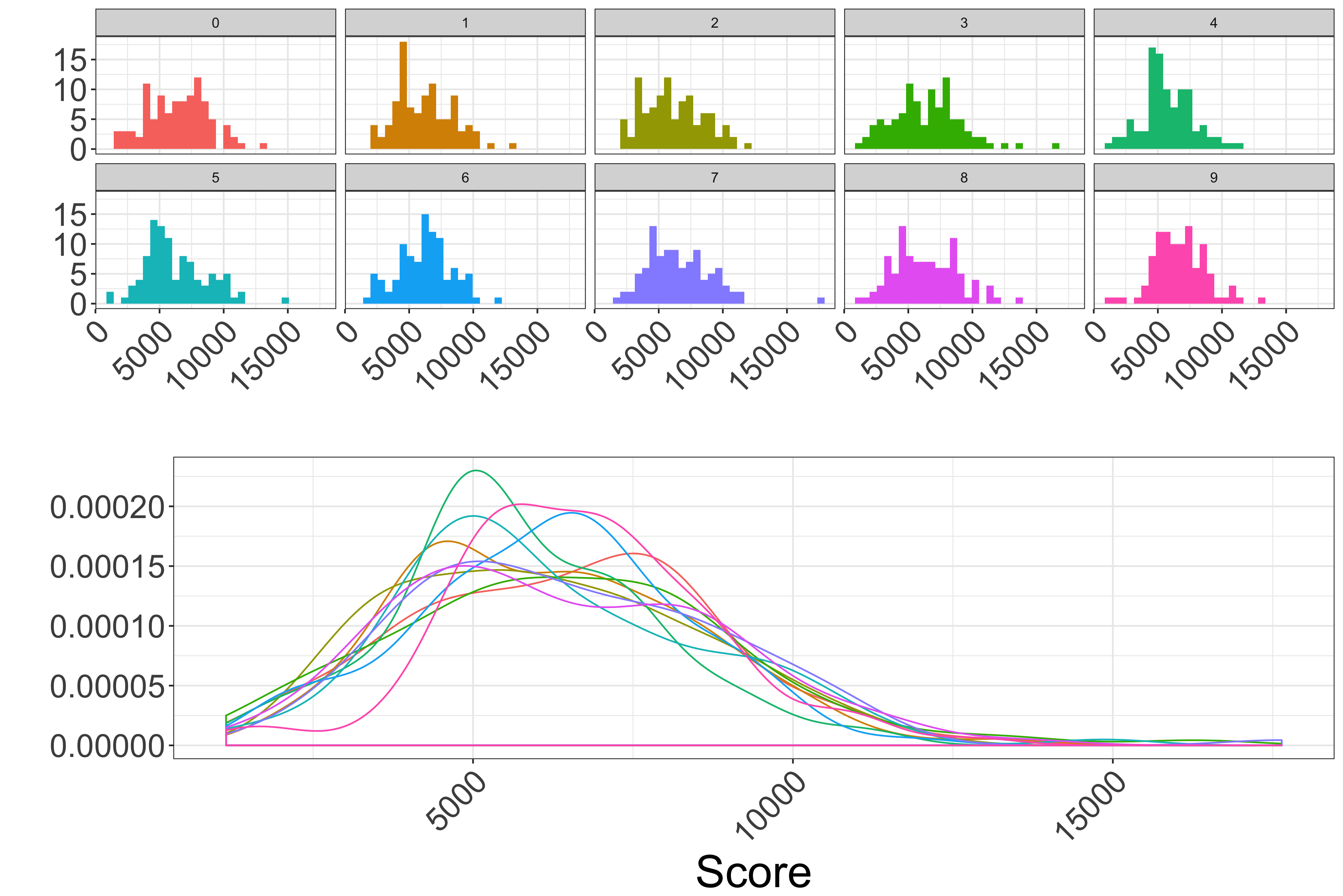}
    \caption{acktr}
    \end{subfigure}%
    \begin{subfigure}[b]{0.33\textwidth}
    \includegraphics[width=0.99\textwidth]{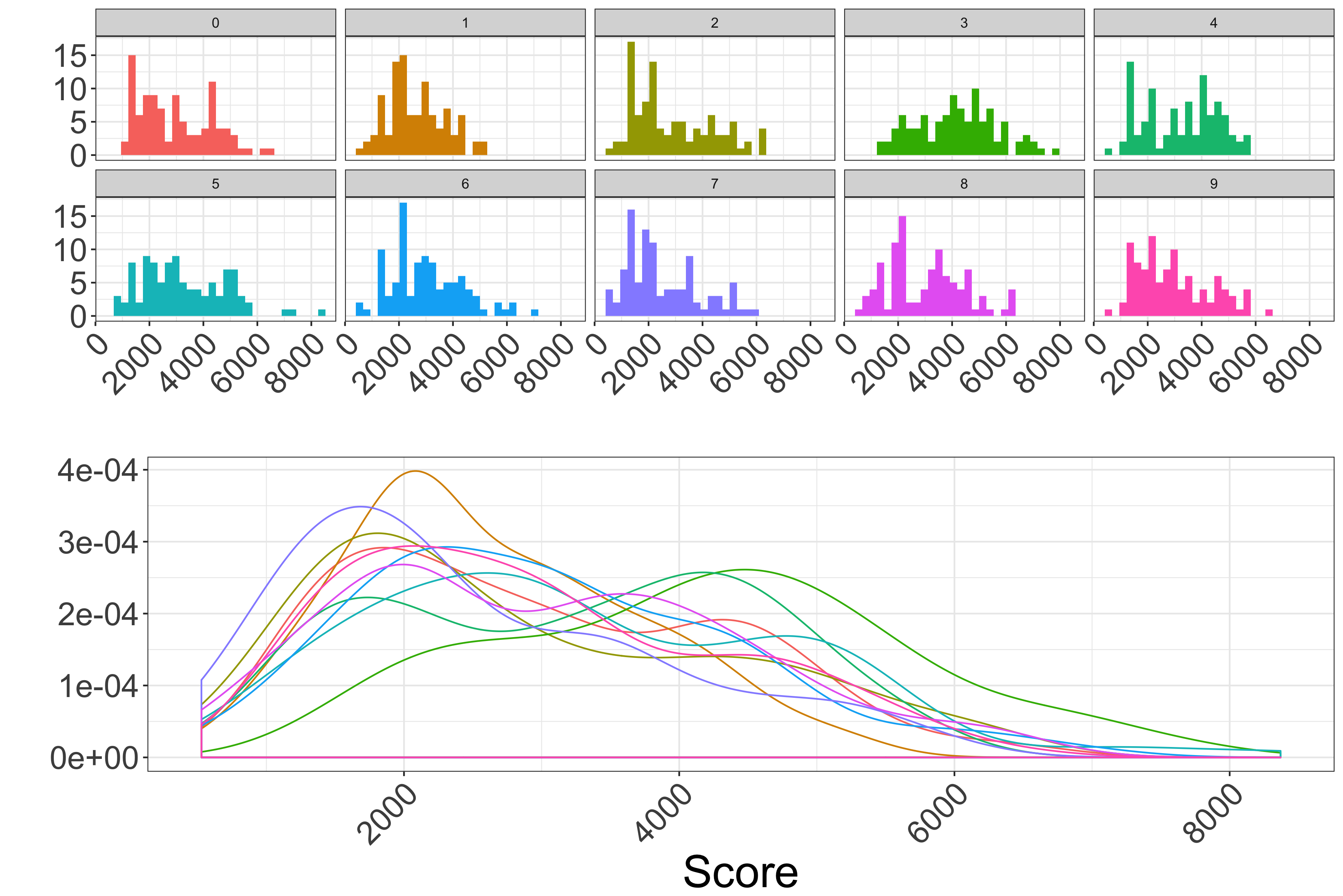}
    \caption{ppo2}
    \end{subfigure}
    \caption{\label{fig:beam-rider-histogram} BeamRider Histograms}
\end{figure}

\begin{figure}[h!]
    \centering
    \begin{subfigure}[b]{0.33\textwidth}
    \includegraphics[width=0.99\textwidth]{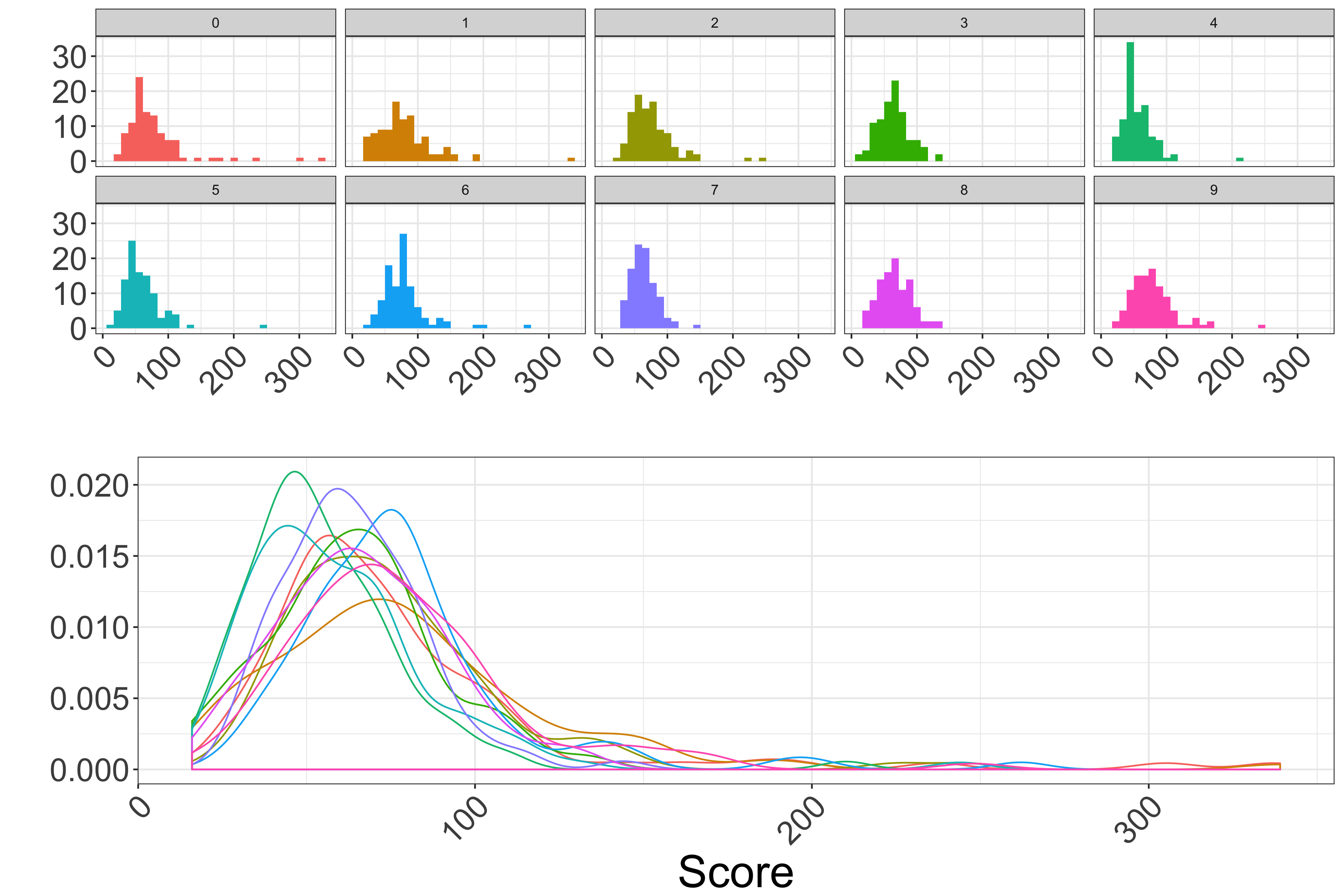}
    \caption{a2c}
    \end{subfigure}%
    \begin{subfigure}[b]{0.33\textwidth}
    \includegraphics[width=0.99\textwidth]{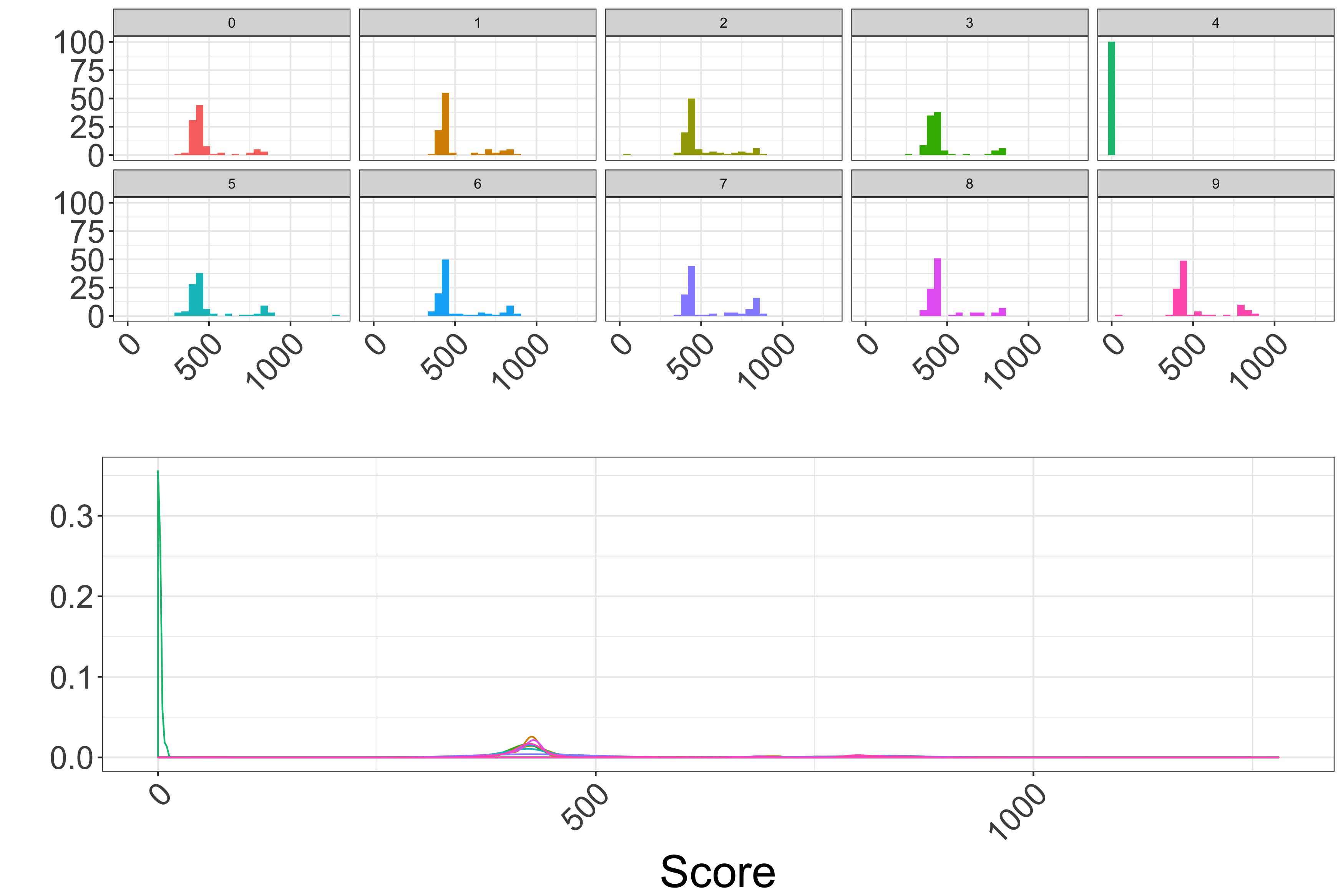}
    \caption{acktr}
    \end{subfigure}%
    \begin{subfigure}[b]{0.33\textwidth}
    \includegraphics[width=0.99\textwidth]{BreakoutNoFrameskip-v4--ppo2_comb.png}
    \caption{ppo2}
    \end{subfigure}
    \caption{\label{fig:breakout-histogram} Breakout Histograms}
\end{figure}

\begin{figure}[h!]
    \centering
    \begin{subfigure}[b]{0.33\textwidth}
        \includegraphics[width=0.99\textwidth]{QbertNoFrameskip-v4--a2c_comb.png}
        \caption{a2c}
    \end{subfigure}%
    \begin{subfigure}[b]{0.33\textwidth}
        \includegraphics[width=0.99\textwidth]{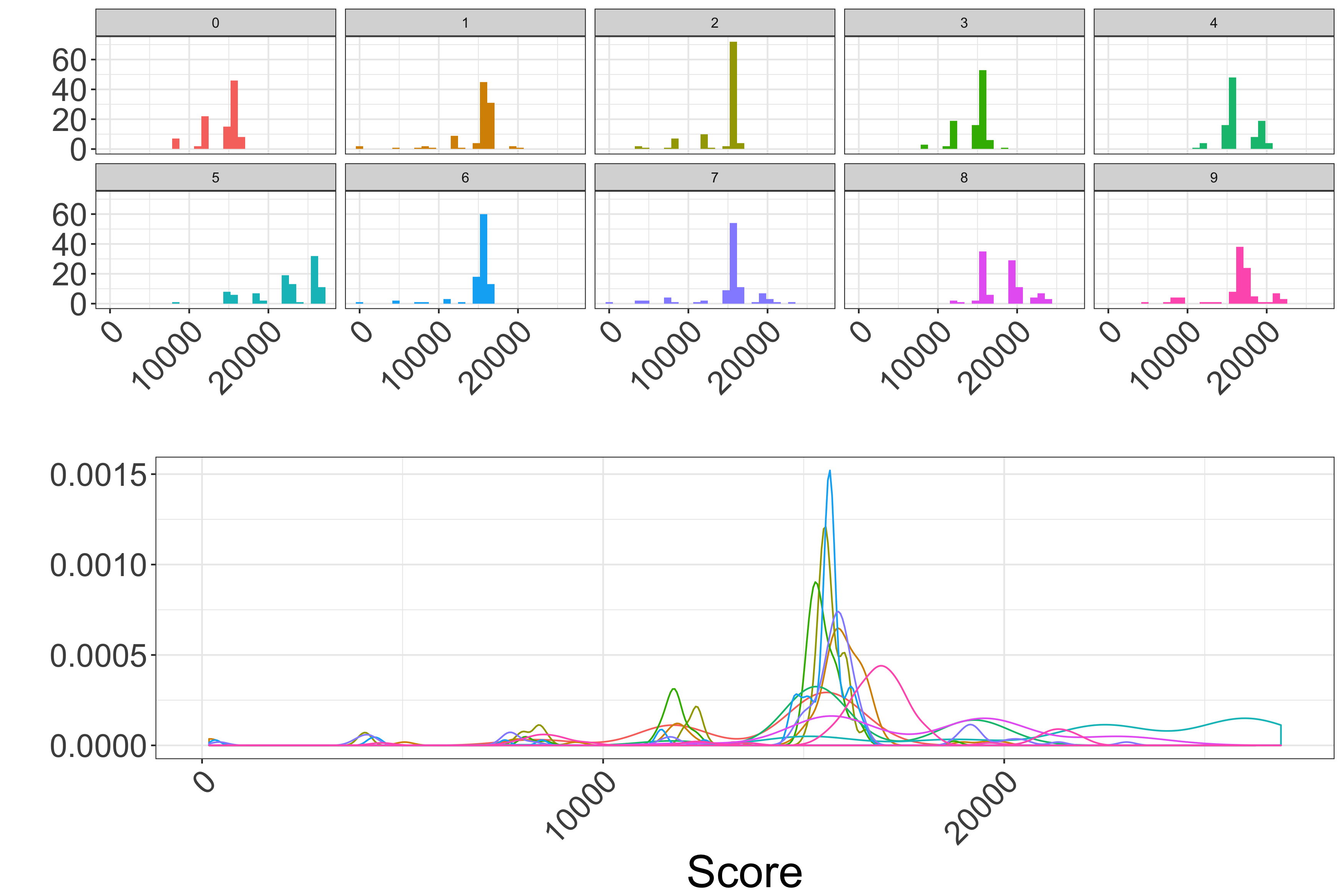}
        \caption{acktr}
    \end{subfigure}%
    \begin{subfigure}[b]{0.33\textwidth}
        \includegraphics[width=0.99\textwidth]{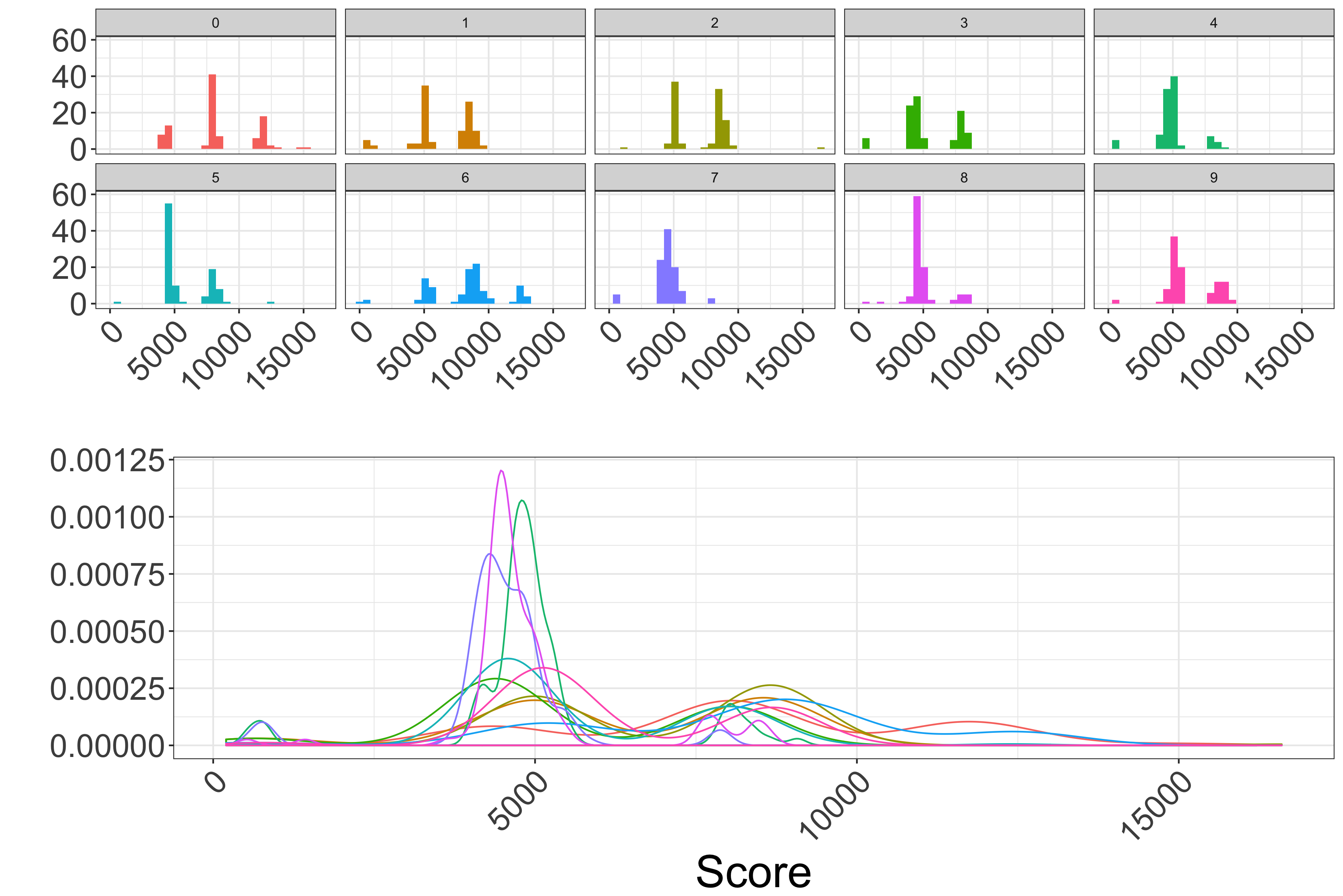}
        \caption{ppo2}
    \end{subfigure}
    \caption{\label{fig:qbert-histogram} Q*Bert Histograms}
\end{figure}

\begin{figure}[h!]
    \centering
    \begin{subfigure}[b]{0.33\textwidth}
        \includegraphics[width=0.99\textwidth]{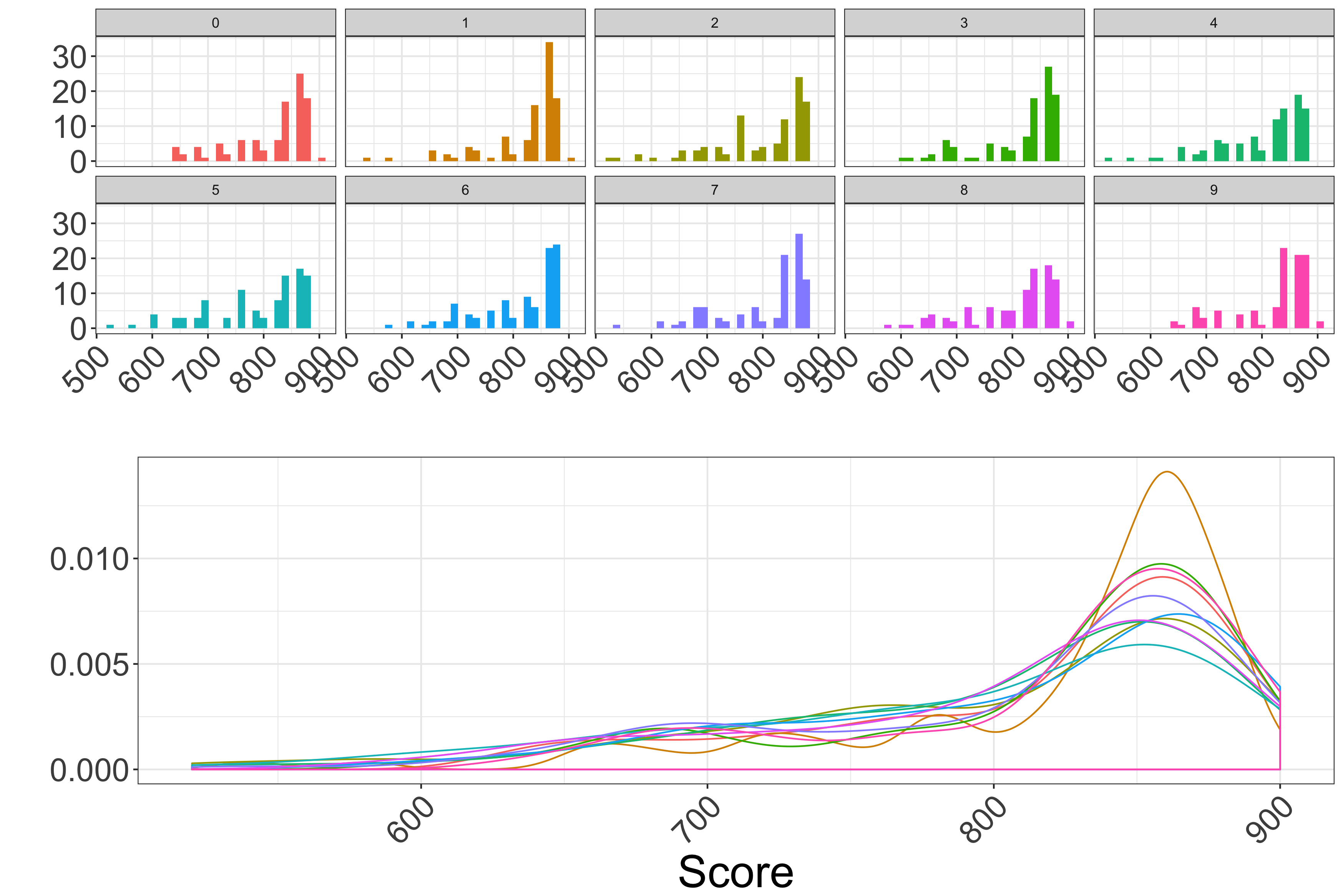}
        \caption{a2c}
    \end{subfigure}%
    \begin{subfigure}[b]{0.33\textwidth}
        \includegraphics[width=0.99\textwidth]{SeaquestNoFrameskip-v4--acktr_comb.png}
        \caption{acktr}
    \end{subfigure}%
    \begin{subfigure}[b]{0.33\textwidth}
        \includegraphics[width=0.99\textwidth]{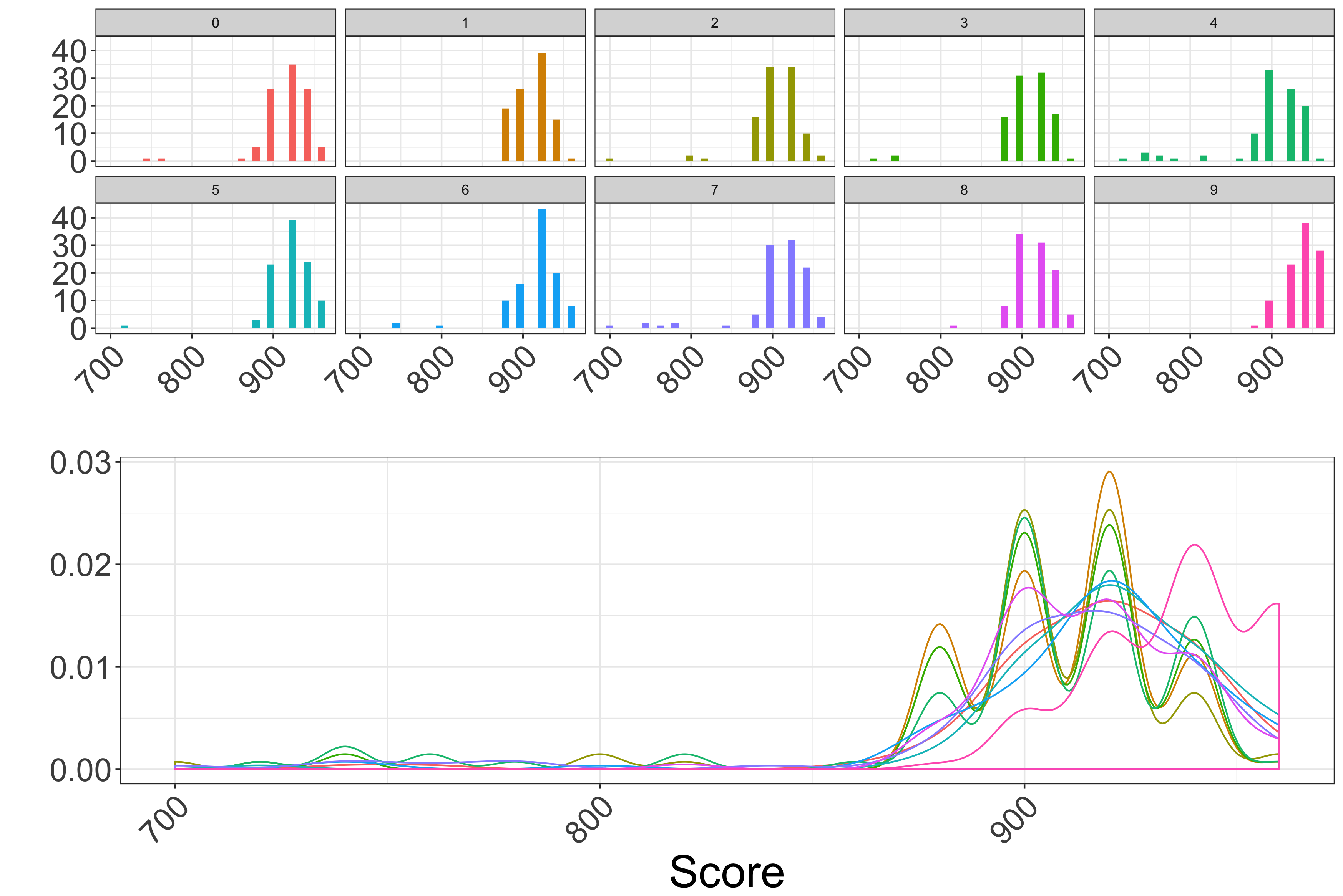}
        \caption{ppo2}
    \end{subfigure}
    \caption{\label{fig:seaquest-histogram} Seaquest Histograms}
\end{figure}

\begin{figure}[h!]
    \centering
    \begin{subfigure}[b]{0.33\textwidth}
    \includegraphics[width=0.99\textwidth]{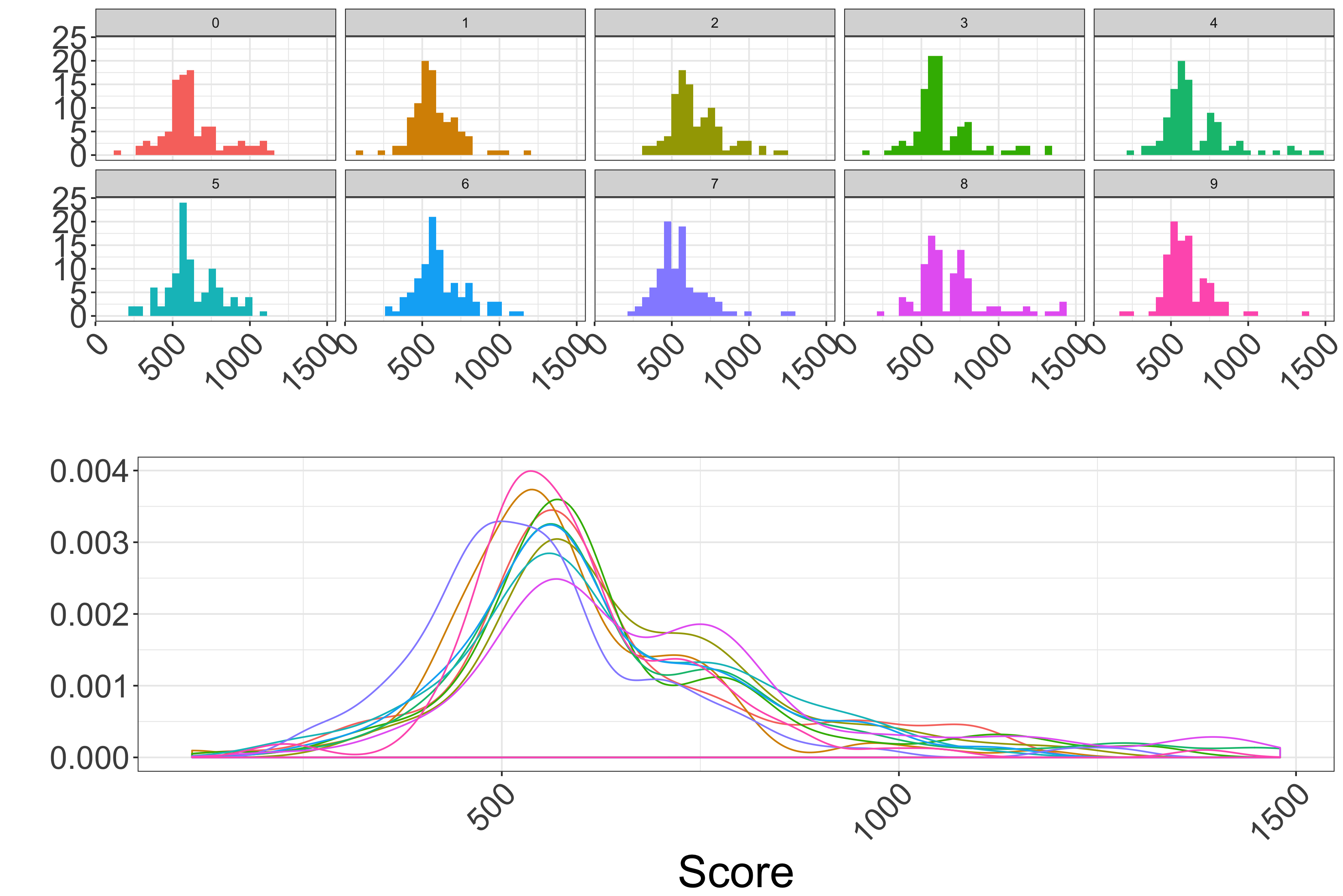}
    \caption{a2c}
    \end{subfigure}%
    \begin{subfigure}[b]{0.33\textwidth}
    \includegraphics[width=0.99\textwidth]{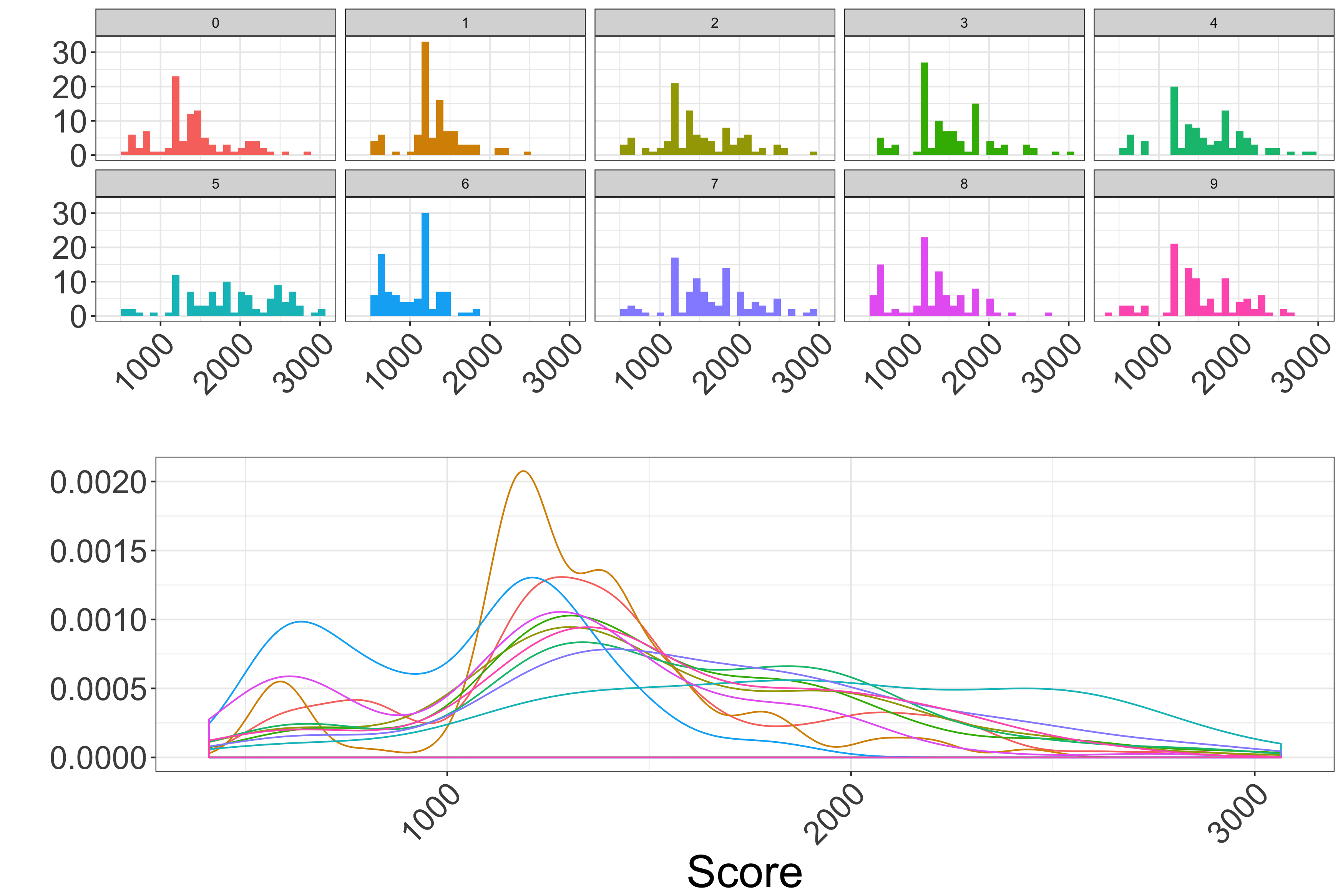}
    \caption{acktr}
    \end{subfigure}%
    \begin{subfigure}[b]{0.33\textwidth}
    \includegraphics[width=0.99\textwidth]{SpaceInvadersNoFrameskip-v4--ppo2_comb.png}
    \caption{ppo2}
    \end{subfigure}
    \caption{SpaceInvaders Histograms}
\end{figure}

\end{document}